\documentclass{article} %
\usepackage{iclr2024_conference,times}

\usepackage{newunicodechar}
\usepackage{booktabs}
\usepackage{graphicx}
\usepackage{enumitem}
\usepackage{multirow}
\usepackage{caption}

\usepackage{amsmath,amsfonts,bm}

\def\eqref#1{equation~\ref{#1}}

\def\1{\bm{1}}

\DeclareMathAlphabet{\mathsfit}{\encodingdefault}{\sfdefault}{m}{sl}
\SetMathAlphabet{\mathsfit}{bold}{\encodingdefault}{\sfdefault}{bx}{n}

\usepackage{hyperref}
\usepackage{url}
\usepackage{alltt}
\usepackage[most]{tcolorbox}
\usepackage{cleveref}
\usepackage{makecell}
\usepackage{tabularx}
\usepackage{longtable}%
\usepackage{arydshln}
\usepackage{fancyvrb}
\usepackage{xspace}
\usepackage{authblk}
\usepackage{tipa} %

\usepackage[nolist]{acronym}

\makeatletter
\newcommand*{\org@overidelabel}{}
\let\org@overridelabel\@verridelabel
\@ifpackagelater{acronym}{2015/03/21}{%
  \renewcommand*{\@verridelabel}[1]{%
    \@bsphack
    \protected@write\@auxout{}{\string\AC@undonewlabel{#1@cref}}%
    \org@overridelabel{#1}%
    \@esphack
  }%
}{%
  \renewcommand*{\@verridelabel}[1]{%
    \@bsphack
    \protected@write\@auxout{}{\string\undonewlabel{#1@cref}}%
    \org@overridelabel{#1}%
    \@esphack
  }%
}
\makeatother

\newcommand{\red}[1]{\textcolor{red}{#1}}
\newcommand{\green}[1]{\textcolor{green}{#1}}
\newcommand{\ours}{\textsc{UMA}\xspace}
\newcommand{\oursfull}{\textsc{UniMedAbstractor (UMA)}\xspace}

\tcbset{
  aibox/.style={
    width=\textwidth,
    top=10pt,
    colback=white,
    colframe=black,
    colbacktitle=black,
    enhanced,
    center,
    attach boxed title to top left={yshift=-0.1in,xshift=0.15in},
    boxed title style={boxrule=0pt,colframe=white,},
  }
}
\newtcolorbox{AIbox}[2][]{aibox,title=#2,#1}

\definecolor{aigold}{RGB}{244,210, 1} 
\definecolor{aired}{RGB}{255,180,181} 
\newunicodechar{≤}{\ensuremath{\leq}}
\newunicodechar{≥}{\ensuremath{\leq}}

\title{Universal Abstraction: \\Harnessing Frontier Models to Structure Real-World Data at Scale}

\author[1*]{Cliff Wong}
\author[1*]{Sam Preston}
\author[1*]{Qianchu Liu}
\author[1]{Zelalem Gero}
\author[1]{Jaspreet Bagga}
\author[1]{Sheng Zhang}
\author[1]{Shrey Jain}
\author[1]{Theodore Zhao}
\author[1]{Yu Gu}
\author[1]{Yanbo Xu}
\author[1]{Sid Kiblawi}
\author[7]{Srinivasan Yegnasubramanian}
\author[7]{Taxiarchis Botsis}
\author[7]{Marvin Borja}
\author[7]{Luis M. Ahumada}
\author[7]{Joseph C. Murray}
\author[2,5]{Guo Hui Gan}
\author[4]{Roshanthi Weerasinghe}
\author[5,6]{Kristina Young}
\author[3,5]{Rom Leidner}
\author[3,5]{Brian Piening}
\author[3,5]{Carlo Bifulco}
\author[1]{Tristan Naumann}
\author[1$\dagger$]{Mu Wei}
\author[1$\dagger$]{Hoifung Poon}

\affil[1]{Microsoft Research, Redmond, WA, USA}
\affil[2]{Providence Portland Medical Center, Portland, OR, USA}
\affil[3]{Providence Genomics, Portland, OR, USA}
\affil[4]{Providence Research Network, Renton, WA, USA}
\affil[5]{Earle A. Chiles Research Institute, Providence Cancer Institute, Portland, OR, USA}
\affil[6]{The Oregon Clinic, Radiation Oncology Division, Portland, OR}
\affil[7]{Johns Hopkins University School of Medicine, Baltimore, MD, USA}
\affil[*]{Equal contributions}

\affil[$\dagger$]{Corresponding authors: muhsin.wei@microsoft.com, hoifung@microsoft.com}

\newcommand{\eat}[1]{\ignorespaces}

\iclrfinalcopy %
\begin{document}

\maketitle

\begin{acronym}
    \acro{AJCC}{American Joint Committee on Cancer}
    \acro{LLM}{large language model}
    \acro{NLP}{natural language processing}
    \acro{RCT}{randomized controlled trial}
    \acro{RWD}{real-world data}
    \acro{RWE}{real-world evidence}
\end{acronym}

\begin{abstract} 

A significant fraction of real-world patient information resides in unstructured clinical text. Medical abstraction extracts and normalizes key structured attributes from free-text clinical notes, which is the prerequisite for a variety of important downstream applications, including registry curation, clinical trial operations, and real-world evidence generation. Prior medical abstraction methods typically resort to building attribute-specific models, each of which requires extensive manual effort such as rule creation or supervised label annotation for the individual attribute, thus limiting scalability. 

In this paper, we show that existing frontier models already possess the ``universal abstraction'' capability for scaling medical abstraction to a wide range of clinical attributes. 
We present \oursfull, a unifying framework for zero-shot medical abstraction with a modular, customisable prompt template and the selection of any frontier \acp{LLM}. 
Given a new attribute for abstraction, users only need to conduct lightweight prompt adaptation in \ours to adjust the specification in natural languages. Compared to traditional methods, \ours eliminates the need for attribute-specific training labels or handcrafted rules, thus substantially reducing the development time and cost.

We conducted a comprehensive evaluation of \ours in oncology using a wide range of marquee attributes representing the cancer patient journey. These include relatively simple attributes typically specified within a single clinical note (e.g., performance status, treatment), as well as complex attributes requiring sophisticated reasoning across multiple notes at various time points (e.g., tumor site, histology, staging). Based on a single frontier model such as GPT-4o, \ours matched or even exceeded the performance of state-of-the-art attribute-specific methods, each of which was tailored to the individual attribute.
To facilitate population-scale real-world data structuring and evidence generation, we will release our code at \url{https://github.com/microsoft/ua_oncophenotype}.

\end{abstract}

\acresetall %

\section{Introduction}

\begin{figure}[t!]
\centering
\includegraphics[width=14cm]{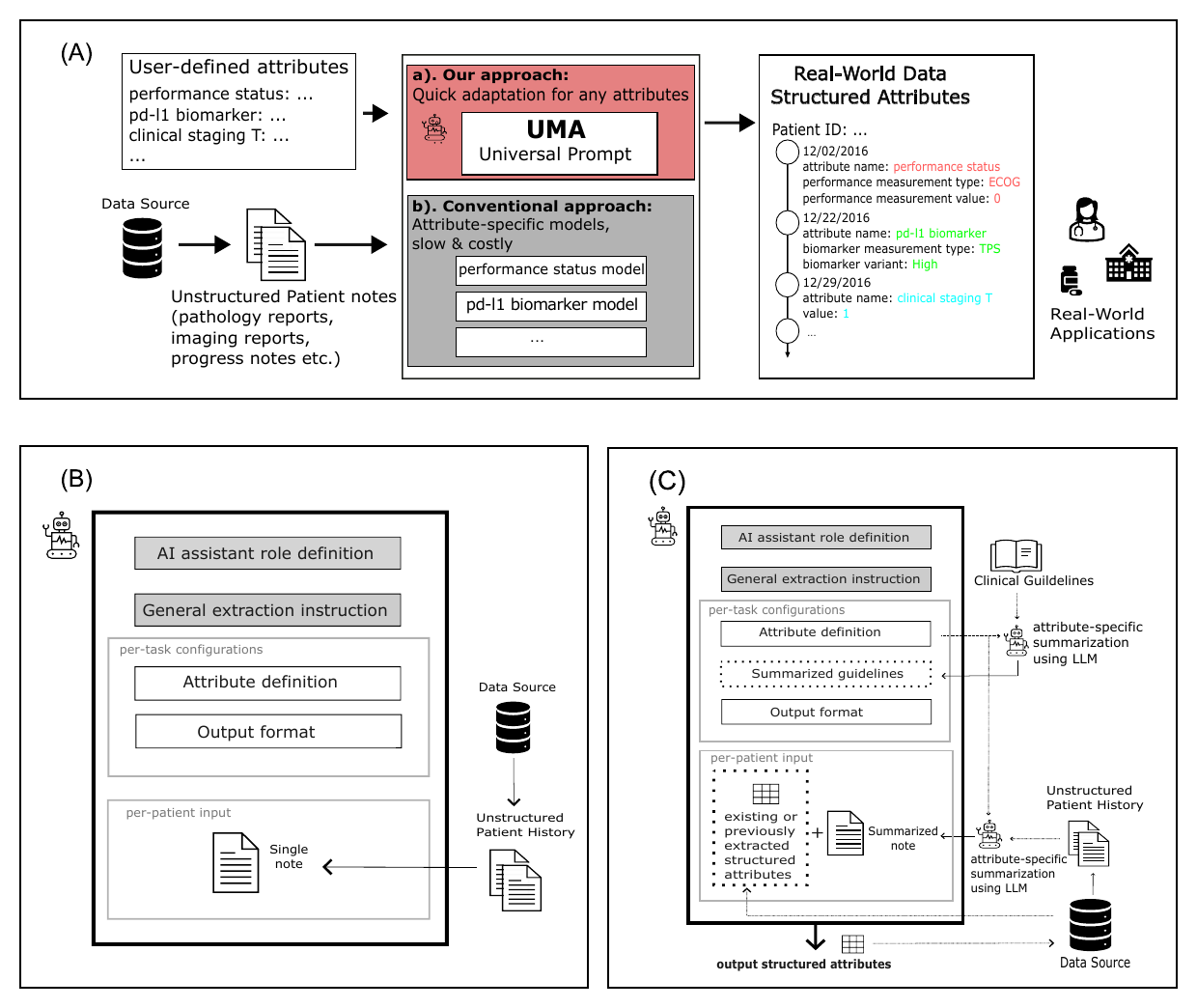}
\caption{(A): Overview of \oursfull, a one-model-for-all universal abstraction method for medical attribute abstraction. In contrast with conventional approaches that build specialized models for each specific attribute (slow and costly due to the manual collection of training labels or heuristic rules), \ours can be quickly configured to abstract any user-defined medical attributes from unstructured patient data in a zero-shot manner. The outcome of the pipeline is structured real-world data that serves as the foundation for real-world applications such as clinical trial matching, data curation for surveillance and monitoring, drug discovery, etc. (B) \ours instantiation template for short-context attributes which can be extracted from within a single
note (an example prompt is in \Cref{fig: example short context prompt}). (C) \ours instantiation template for long-context attributes (e.g. cancer staging attributes) which require complex reasoning over multiple notes and long clinical guidelines (an example prompt is in \Cref{fig: example long context prompt}). We also provide the flexibility to
chain the prompts to leverage previously extracted attributes.}
\label{fig:uma_overview}
\end{figure}

\Ac{RWD} in healthcare refers to information collected from records representing standard medical care as opposed to data from research clinical trials. 
\Ac{RWD} can offer a more comprehensive view of patient experiences, helps optimize healthcare delivery, and supports more informed decision-making across the healthcare ecosystem. Particularly, \ac{RWD} can be used to generate \ac{RWE} which is increasingly utilized for medical evidence generation, providing a complement to the existing standard use of a \ac{RCT}. A significant proportion of \ac{RWD} comes from unstructured patient data, such as dictated progress notes and radiology reports, which store much of the patient information needed to improve care and clinical research. Medical abstraction is a process that structures \ac{RWD} by extracting and normalizing information from unstructured patient records. Traditional medical abstraction methods for structuring \ac{RWD} still require substantial manual efforts, such as crafting extraction rules or annotating examples for supervised learning. These manual efforts are expensive and time-consuming. In the U.S. alone, there are close to two million new cancer patients each year and curating key information for a single patient takes hours. As such, traditional abstraction is a significant barrier to realizing the full opportunity presented by the digitization of medical records~\citep{rudrapatna2020opportunities}. 
Moreover, evolving guidelines for defining key cancer attributes further compounds the challenge, leading to semantic drift and therefore often rendering previously collected labels outdated~\cite{gao2021limitations,gao2019classifying,preston2023toward}.

While traditional methods face challenges in efficiently scaling up medical abstraction due to the need to develop specialized models for each specific attribute, state-of-the-art \acp{LLM} such as GPT-4o have demonstrated emergent capabilities in biomedical applications without requiring any specialized training~\cite{lee2023benefits, nori2023can}. 
In this paper, we propose \oursfull, a framework that harnesses the universal structuring capabilities of \acp{LLM} for zero-shot universal abstraction. Here we define universal abstraction as a one-model-for-all approach that can efficiently scale to extract new attributes from any ontology, adapt to evolving guidelines, and handle new patients and datasets across different institutions. The key to \ours is a modular template that can flexibly incorporate patient data and the user-defined task definitions for each attribute. \ours can also accommodate different types of input (single note, multiple notes, long guidelines, etc.) and requirements for abstracting different types of attributes (\Cref{fig:uma_overview}A).

We test \ours in oncology, where medical abstraction is particularly challenging. Our main experiments are conducted on real-world data from the Providence Health System, a large integrated delivery network. 
Without requiring any specialized training, \ours displays impressive universal abstraction capabilities. \ours with GPT-4o surpasses conventional baselines in the overall performance by around 2 absolute points in F1/accuracy scores. Its performance further improves on the more challenging long-context attribute abstraction tasks when using the advanced reasoning model O1.
In some cases, such as pathologic T in cancer staging, \ours even outperforms the supervised method by over 20 absolute points in accuracy. 
We also demonstrate that \ours generalizes more effectively than supervised models. Evaluated on a comparable real-world dataset from another institution (Johns Hopkins), \ours shows greater robustness to distributional shifts. Additionally, \ours achieves a substantial reduction in cost and human labor, cutting both to less than one-tenth of the conventional approach.

\eat{
Through this exploration we make the following contributions:
\begin{itemize}
    \item We evaluate the performance of recent large language models on the extraction of fifteen key attributes from patient records, a level of performance that suggests to a universal structuring capacity and may enable the creation of fit-for-purpose structured patient representations.
    \item We demonstrate that GPT-4 can effectively abstract attributes by ingesting relevant standard guidelines provided in the prompt, enabling rapid development of abstraction processes to adapt to new guideline versions or new attributes without additional supervised training.
    \item We provide a thorough ablation study on prompt configuration and identify best practices in handling challenges that typify clinical text, such as context length limits.
\end{itemize}
}

\section{Universal Abstraction with \ours}

We propose \ours (pronounced as \textipa{~["u:ma"]}), a universal abstraction framework to leverage large language models to structure real world data from unstructured patient notes in a zero-shot manner. \ours uses a flexible prompt template that allows users to specify target attributes, combining predefined components, task-specific configurations, and per-patient inputs. It also includes advanced modules to support complex, long-context attributes requiring long-term reasoning. An overview of the template is shown in \Cref{fig:uma_overview}B and \Cref{fig:uma_overview}C.

\subsection{Pre-defined: general task introduction}
We begin by providing the \ac{LLM} with a general understanding of the medical abstraction task, enabling broad applicability across attributes. The universal prompt template starts with predefined, generic instructions that frame the task as event extraction, where each event group corresponds to an attribute occurrence in the patient record. Each group contains specific descriptors that characterize the event. This setup allows users to later specify attribute definitions, relevant descriptors, and expected outputs for each occurrence.

\subsection{Attribute-specific task configurations}

We modularize the task configurations to ensure scalability for new attributes. For each attribute, we configure the following components:

\paragraph{Attribute definition} For each task, we set up an attribute definition block in the prompt where the user can specify the attribute's meaning, relevant descriptors, and expected output formats—e.g., verbatim text or predefined categories.
For example, the performance status attribute may include:

\begin{verbatim}
"performance status measurement type": 
    name of measurement type: ECOG, KPS, PPS, or Lansky
"performance status value":
    the measured value of performance status measurement type.
    Extract only the numerical value.

\end{verbatim}

To improve interpretability and accuracy, users can also define contextual descriptors (e.g., note date, certainty, source spans, model reasoning), which aid both postprocessing and guiding the \ac{LLM} toward correct outputs (see ablation results in \Cref{fig:Ablation}).

\paragraph{Output format} 
To streamline extraction and post-processing, outputs are formatted as a list of JSON objects, with each representing an attribute occurrence. We provide an output template to guide the \ac{LLM}, where each dictionary maps descriptor names to values extracted from the text. For example, the output format for performance status is:

\begin{verbatim}
[
  {
  "performance status measurement type":
  <performance status type>,
  "performance status value":
  <performance status value>
  }
  ...
  ]
\end{verbatim}

\paragraph{Long-context Attribute: Incorporating clinical guidelines}
Some tasks require integrating complex clinical guidelines, such as those from the ICD-O (240+ pages) or AJCC Cancer Staging Manual (600+ pages). Due to \ac{LLM} context limitations, directly including these guidelines is impractical. To address this, we perform a one-time structuring of the guidelines using GPT-4 to generate attribute-specific summaries. By incorporating the attribute definition block into the summarization prompt, we ensure relevance and efficiency in guiding the model (see \Cref{fig:uma_overview}C).

\subsection{Patient Input}
Once the prompt template is configured, patient data—typically unstructured text from sources like pathology, imaging, surgical, or progress notes—is provided as input. The template is designed to flexibly accommodate any note type.

\paragraph{Long-context Attributes: Reasoning Across Patient History}
For tasks requiring longitudinal reasoning, inserting individual notes is insufficient. Patients often have 20+ notes spanning their clinical history. To handle this, we apply GPT-4 to perform attribute-specific summarization across notes, generating a concise, chronological patient summary (see \Cref{fig:uma_overview}C).

This approach reduces input length, cuts computational costs and latency, and improves performance by eliminating duplicated or irrelevant content—long known to impair \ac{LLM} quality~\cite{searle2021estimating,liu2024lost,mirzadeh2024gsm}. We demonstrate the impact of this summarization in our ablation studies (\Cref{fig:Ablation}).

\paragraph{Long-context Attribute: Leveraging Structured Data}
The prompt template includes an optional block to incorporate existing or previously extracted structured data alongside unstructured input. This contextual information can narrow the focus of extraction and improve accuracy. For example, knowing the treatment date helps restrict clinical staging to pre-treatment notes, and providing the tumor site can aid staging attribute extraction.

\paragraph{Postprocessing}

Thanks to its modular design, integrating \ours into the attribute abstraction pipeline is seamless. After obtaining the JSON output from the \ac{LLM}, we apply post-processing to refine and finalize the extracted attributes. Descriptors in the output can be used to filter for high-quality occurrences—for example, excluding non-cancerous events based on a disease type descriptor improves precision in response extraction.

Extracted values are then normalized to standard medical ontologies (e.g., ICD-10-CM, NCI Thesaurus, HGNC, HGVS) or to numerical lab values. We use heuristics to expand synonyms for each entity and apply string matching for normalization.

To reduce redundancy, duplicate values across or within notes with similar timestamps are merged into a single attribute group. Each occurrence is then linked to the patient ID and note timestamp where applicable, enabling timeline construction for downstream applications.

\section{Tasks and Experiment Setup}

As a testbed, we apply \ours to abstract a representative set of oncology attributes, spanning both short-context attributes—extractable within a single note—and long-context attributes that require reasoning over longitudinal records and complex clinical guidelines. Short-context attributes include PD-L1, performance status, treatment, response, progression, and case finding. Long-context attributes include coarse- and fine-grained primary site, histology, and clinical/pathologic T, N, and M stages. A full list of attributes and their associated descriptors is provided in Table S1.

Datasets were drawn from two health systems—Johns Hopkins and Providence—and comprise diverse patient documents. Attribute-specific datasets vary based on the availability of manual ground-truth annotations (see Tables S2 and S3). Long-context labels were obtained from cancer registries, while short-context labels were curated in collaboration with Providence clinicians.

All \ours experiments are conducted in a zero-shot setting without training labels. For optimal performance, we use state-of-the-art \acp{LLM}, including GPT-4 and GPT-4o (version: 2024-05-13). For long-context tasks requiring complex reasoning, we additionally evaluate O1 (version: 2024-12-17). Where training labels exist, we compare against BERT-based supervised baselines. In their absence, we report heuristic baselines based on domain-specific rules from \citet{gonzalez2023trialscope}. We report F1 scores for short-context attributes and accuracy for long-context attributes.

Further dataset and experimental details are available in \Cref{appendix: short context attributes} and \Cref{appendix: long context attribute}.

\eat{
\green{ask sam or cliff: do we need to change anything here? ask Tristan for Providence ID. Sam is following up with Hopkins to IRB code}
This work was performed under the auspices of the independent institutional review board (IRB)-approved research protocols (Providence protocol ID \red{2019000204} \eat{; Johns Hopkins Medicine \red{0000000000})} and
was conducted in compliance with human subjects research and clinical
data management procedures—as well as cloud information security policies
and controls—administered within each institution. All study
data were integrated, managed, and analyzed exclusively and solely on respective institute cloud infrastructures. All study personnel completed and
were credentialed in training modules covering human subjects research,
use of clinical data in research, and appropriate use of IT resources and
IRB-approved data assets.}

\section{Experimental Results}

\subsection{Short-context Attribute Abstraction}

As shown in the top sub-figure in \Cref{fig:uma_main_results}, \ours with GPT-4o delivers the best overall performance across six short-context attribute abstraction tasks, achieving the best F1 score. We also observe an overall performance gain when moving from GPT-4 to GPT-4o, demonstrating that \ours effectively leverages improvements in base \acp{LLM}.
Notably, for the PD-L1 biomarker and performance status abstraction tasks, \ours with GPT-4o already reaches ceiling performance. Across the individual tasks, \ours with GPT-4o either outperforms or matches the baseline apart from case finding where the supervised baseline is trained on tens of thousands of labels. It is worth highlighting that the human effort required to develop the case finding model far exceeds that of \ours. While the baseline model relies on over 10,000 hours of manual data curation, \ours requires less than an hour to define the attributes via prompt engineering. The strong overall performance and substantial cost savings of \ours make it a compelling alternative to traditional attribute extraction methods.

\subsection{Long-context Attribute Abstraction: Cancer Staging}

\begin{figure}[h]
\centering
\includegraphics[width=0.8\textwidth]{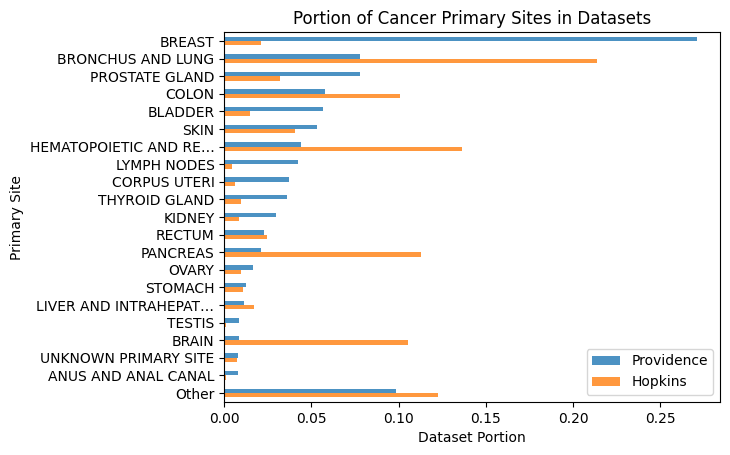}
\caption*{Figure S1. Distribution of primary sites in Providence and Johns Hopkins datasets.}
\label{fig:supplementary_primary_site_dist}
\end{figure}

The second sub-figures in \Cref{fig:supplementary_primary_site_dist} compare \ours and supervised baselines on long-context cancer staging abstraction tasks in the Providence dataset. Despite operating in a zero-shot setting\footnote{While full examples are not provided, some prompts include fixed snippets to clarify guideline application.}, \ours outperforms the supervised model by over 2 absolute points in accuracy, including a 20-point gain on the pathologic T attribute. Similar to short-context tasks, \ours benefits from advances in the underlying model—particularly with O1—showing notable improvements on complex, reasoning-heavy TNM attributes.

While \ours lags behind the supervised baseline on certain tasks such as fine-grained primary site, histology, and clinical T, its performance remains impressive given the setting. The supervised models were trained on split subsets of the same dataset (i.e. allowing them to learn dataset-specific annotation nuances) and required over 10,000 hours of manual labeling. In contrast, \ours relies on under an hour of prompt design, highlighting its efficiency and scalability.

\begin{figure}[ht]
\centering
\includegraphics[width=15cm]{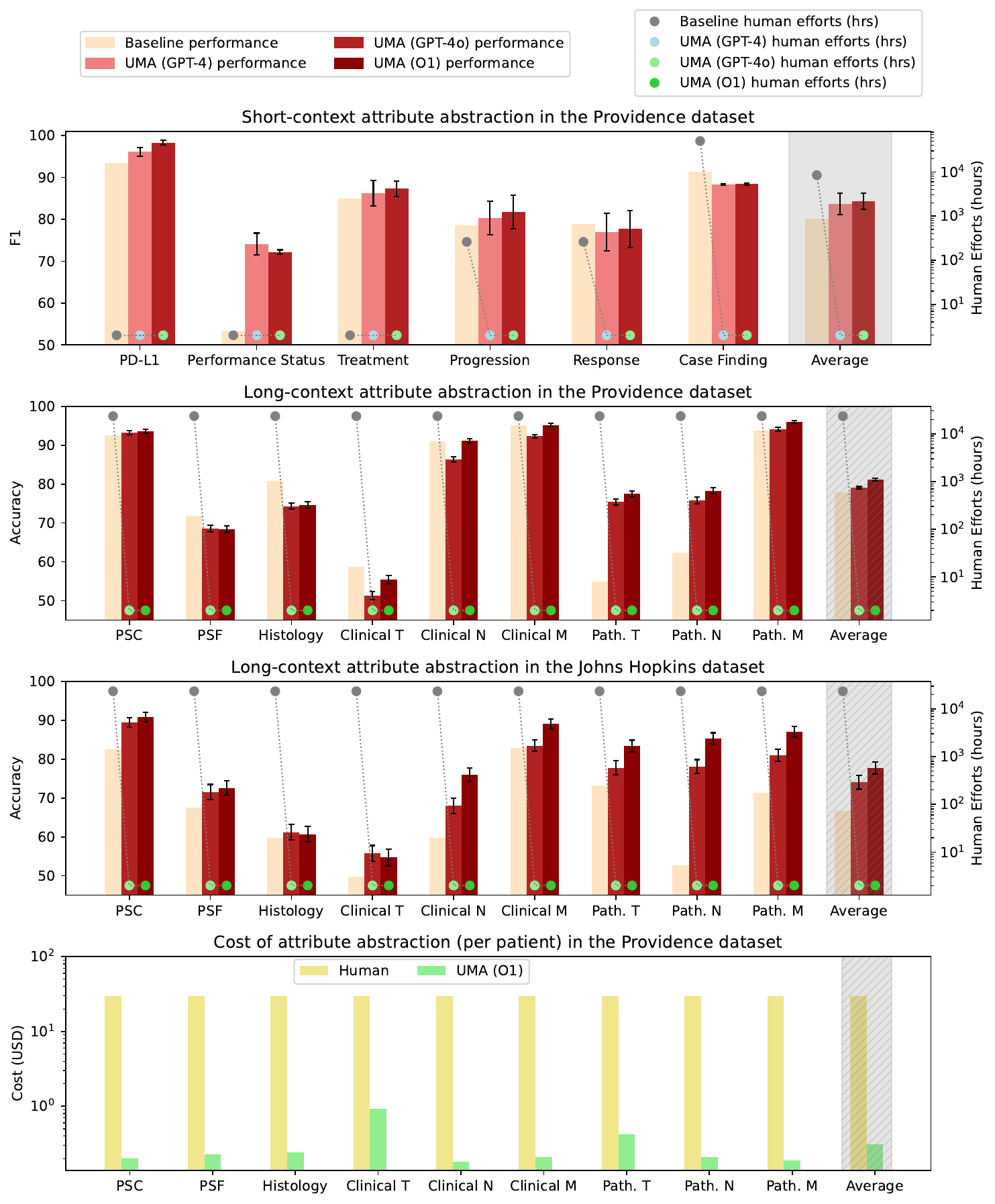}
\caption{\ours main results. The top three sub-figures represent the performance on short-context attribute abstraction in Providence data and long-context attribute abstraction across both Providence and Johns Hopkins datasets. Standard deviation from 1000 bootstrapping sampling procedures is reported
 as error bars. Detail numbers are in Table S4, Table S5, and Table S6. All the baseline approaches are supervised baselines except for PD-L1, Performance Status and Treatment where the baseline approach is heuristics-based as no training data is available. The number of human hours involved are estimated from 1 hour per case as reported in \citep{miller2024cancer}. The last row represents the inference cost for \ours (O1) (also see Table S7) as compared with the cost of hiring humans to annotate the same testsets. Human annotation cost is approximated using the average hourly wage of a cancer registrar (30 USD/hour) \citep{ncra_salary2022}. We show that \ours achieves an overall better performance across different attribute types and across different institutions with significant cost and human effort reduction compared with conventional baseline approaches for medical abstraction.}
\label{fig:uma_main_results}
\end{figure}

A further limitation of supervised models is their dependence on labeled data, which is rarely available across all institutions. The third sub-figure in \Cref{fig:uma_main_results} evaluates model generalizability by applying the Providence-trained supervised model and \ours to the Johns Hopkins dataset, which was only used for evaluation. Here, \ours significantly outperforms the supervised baseline, demonstrating greater robustness to distribution shifts and superior generalization. Unlike supervised models that risk overfitting to spurious correlations in training data, \ours relies solely on structured prompts grounded in clinical guidelines, making it a more reliable solution for deployment across diverse or unseen datasets.

\begin{figure}[t!]
\centering
\includegraphics[width=14cm]{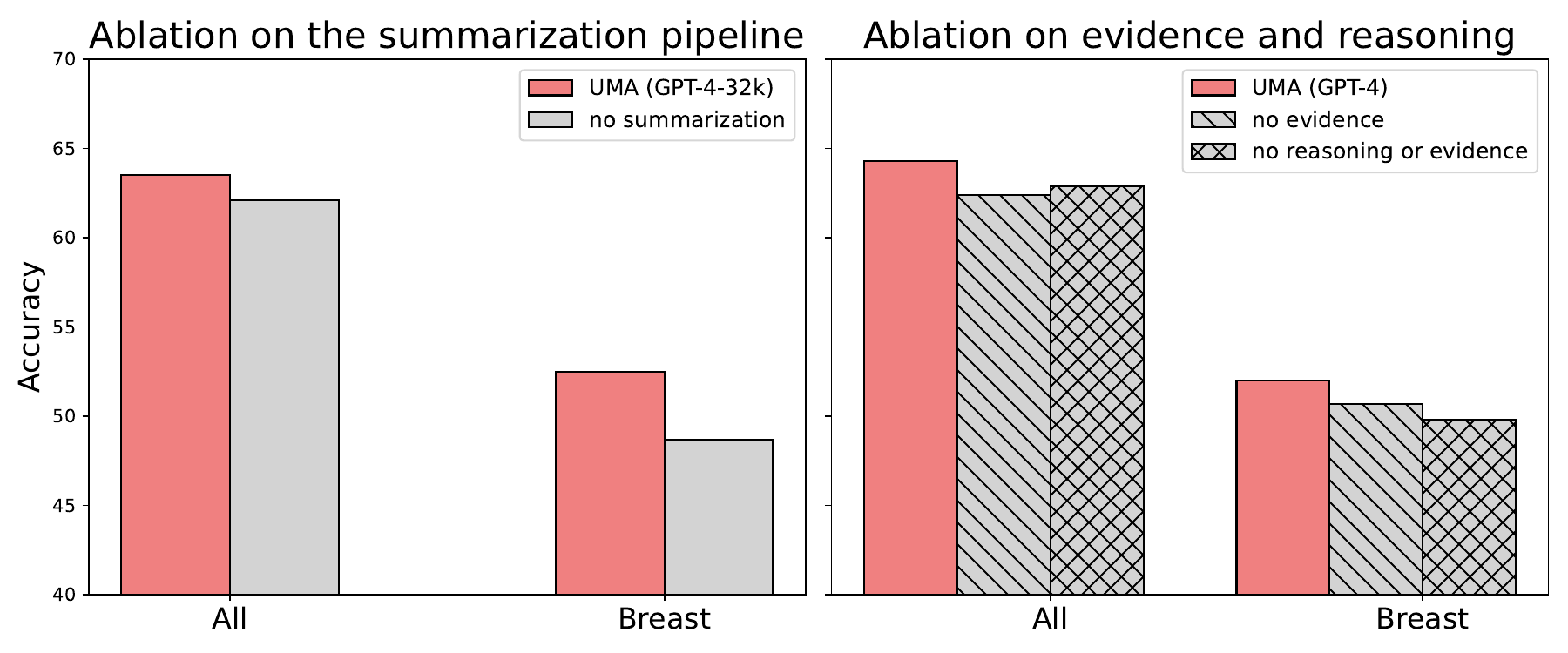}
\caption{Ablating the key components (summarization, evidence, reasoning) in \ours for abstracting long-context attributes (fine-grained primary site) on the Providence dataset. ``All'' includes all tumor sites, while ``Breast'' includes only the more challenging abstraction tasks for `breast' cancer patients. \label{fig:Ablation}}
\end{figure}

To assess the contribution of key components in \ours, we conducted ablation studies on the summarization step and the use of evidence and reasoning descriptors in the attribute definition block (\Cref{fig:Ablation}). While summarization primarily addresses context length limitations, we evaluated its impact on performance using GPT-4-32k, which has a larger context window. Comparing two setups—one with summarization and one using concatenated full notes (restricted to patients within the 32k-token limit)—we found that summarization not only improves efficiency but also enhances accuracy, likely by filtering out irrelevant information. We also ablated the attribute definition prompt by removing the evidence and reasoning descriptors. This led to a performance drop, particularly on the breast cancer dataset, where abstraction tasks are more complex. These descriptors serve as grounding mechanisms, akin to chain-of-thought reasoning, and improve model accuracy on tasks requiring deeper inference.

\section{Discussion}

\subsection{The Potential for \ours Integration into Clinical Workflows}

The strong performance and rapid development cycle of \ours make it highly scalable for universal abstraction. Unlike conventional methods that require months or years to curate training data or craft heuristics, \ours enables near-instant onboarding of new attributes—limited only by the time needed to define them in prompts. This allows for fast, large-scale generation of structured patient records to support a wide range of downstream applications, from clinical decision support to trial recruitment. Moreover, \ours can easily accommodate evolving clinical guidelines, a critical feature as many now follow rolling updates. In cancer staging, for instance, different guideline versions must be applied based on diagnosis dates, a requirement \ours can seamlessly handle.

Despite \ours's strong performance, clinical use still requires human oversight. \ours facilitates efficient review by producing interpretable outputs: summarized patient histories for context, evidence descriptors for traceability, and reasoning descriptors that explain model logic—serving as a built-in chain-of-thought. Additionally, \ours is well-suited for long-context abstraction tasks, where efficiency is paramount. Summarization not only improves token usage but also accelerates human review. In dynamic clinical settings, updated abstractions can be generated by summarizing new notes and appending them to existing summaries, enabling continuous, efficient updates to structured patient data.

\subsection{Limitations}

A key limitation of our study is data quality, particularly missing information due to the fragmented nature of patient records. Unlike cancer registries that aggregate data across providers, our dataset is confined to a single EHR system, limiting visibility into a patient’s full medical history. Future work should incorporate notes from multiple hospital systems and clinics to improve the completeness and accuracy of longitudinal patient data.

Additionally, this study serves as a proof-of-concept, showcasing the effectiveness of basic prompting with LLMs. While our simple approach already matches or outperforms conventional methods, we did not explore more advanced prompting strategies—such as self-verification, self-consistency, or few-shot learning—which could further improve performance. Evaluation of a broader range of \acp{LLM}, including open-source models like DeepSeek R1 and LLaMA, is also left for future work.

\section{Related Work}

\subsection{Conventional approaches to automate medical abstraction}

Automated data extraction using NLP and machine learning has been widely explored in oncology. \citet{Gauthier2022-wk} demonstrated that automated extraction from EHRs of advanced lung cancer patients was both accurate and faster than manual abstraction, despite challenges like unstructured text and non-standard terminology. \citet{Preston2023-wb} used registry-derived, patient-level supervision to train deep neural network models for cross-document extraction, achieving high performance on core tumor attributes and highlighting potential for accelerating registry curation. \citet{kefeli2024generalizable} proposed a BERT-based model to classify TNM stages directly from pathology reports, using publicly available data.

\subsection{LLMs in medical abstraction}
Recent studies have applied LLMs like GPT-4 to abstract clinical attributes from oncology notes, showing strong performance in tasks such as Named Entity Recognition and relation extraction \cite{zhou2024universalner, bhattarai2024leveraging, goel2023llms, 10.1093/jamia/ocad259, wong2023scaling}. These approaches typically use prompt engineering and few-shot learning to identify entity spans, types, and relationships. However, existing work lacks scalable prompting strategies for handling multiple attributes and end-to-end evaluation. Furthermore, little attention has been given to guideline-based classification and evidence aggregation across notes—gaps our study aims to address.

\section{Conclusion}
In this paper, we introduced \ours, a zero-shot and one-model-for-all framework that utilizes an underlying \ac{LLM} to automate medical abstraction across multiple attributes from unstructured clinical notes. Through a flexible universal prompt template, \ours achieves universal abstraction in the way that it can easily generalize to different types of attributes (including both simple short-context and complex long-context oncology attributes) and can cope with long input, complex reasoning and involving guidelines. Compared with the conventional approaches that build attribute-specific models, \ours is a single model that is both more scalable with much lower adaptation costs for onboarding new attributes and achieving better overall performance. In particular, \ours improves generalizes significantly better than supervised baselines across different datasets from different institutions. We believe that \ours provides a promising direction for medical abstraction in the future with great potential to enhance the efficiency, scalability and usability in the clinical workflow.

\section{Human subjects/IRB, data security, and patient privacy}
This work was performed under institutional review board (IRB)-approved research protocols (Providence protocol ID 2019000204, Johns Hopkins protocol ID IRB00482075) and was conducted in compliance with human subjects research and clinical data management procedures—as well as cloud information security policies and controls—administered within Providence Health and Johns Hopkins Medicine. All study data were integrated, managed, and analyzed exclusively and solely on Providence-or Johns Hopkins-managed cloud infrastructure. All study personnel completed and were credentialed in training modules covering human subjects research, use of clinical data in research, and appropriate use of IT resources and IRB-approved data assets.

\section{Notes}

We thank the inHealth Precision Medicine program, the Armstrong Institute Cancer Registry team, and the Sidney Kimmel Comprehensive Cancer Center Precision Medicine and Molecular Tumor Board groups at Johns Hopkins for their help and support in curating the Johns Hopkins dataset and providing guidance on evaluation of model performance. We also thank Providence Portland Medical Center, Providence Genomics, the Providence Research Network, the Providence Cancer Institute, and the Oregon Clinic for their assistance in curating and labeling the Providence dataset and guiding the evaluation process.

\begin{figure}[t!]
\centering
\includegraphics[width=14cm]{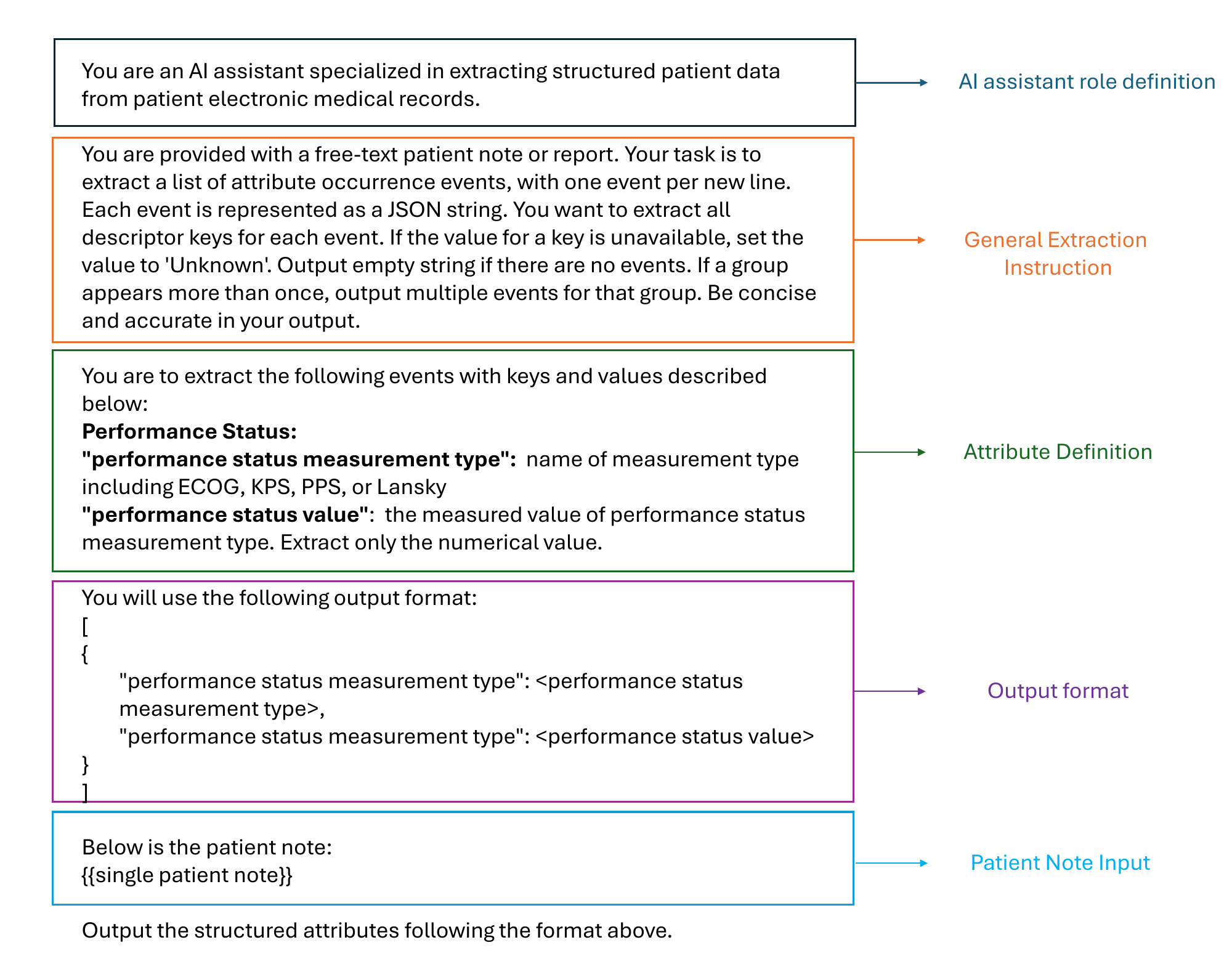}
\caption{Example short-context attribute \ours prompt for extracting performance status from patient note \label{fig: example short context prompt}}
\end{figure}

\begin{figure}[t!]
\centering
\includegraphics[width=14cm]{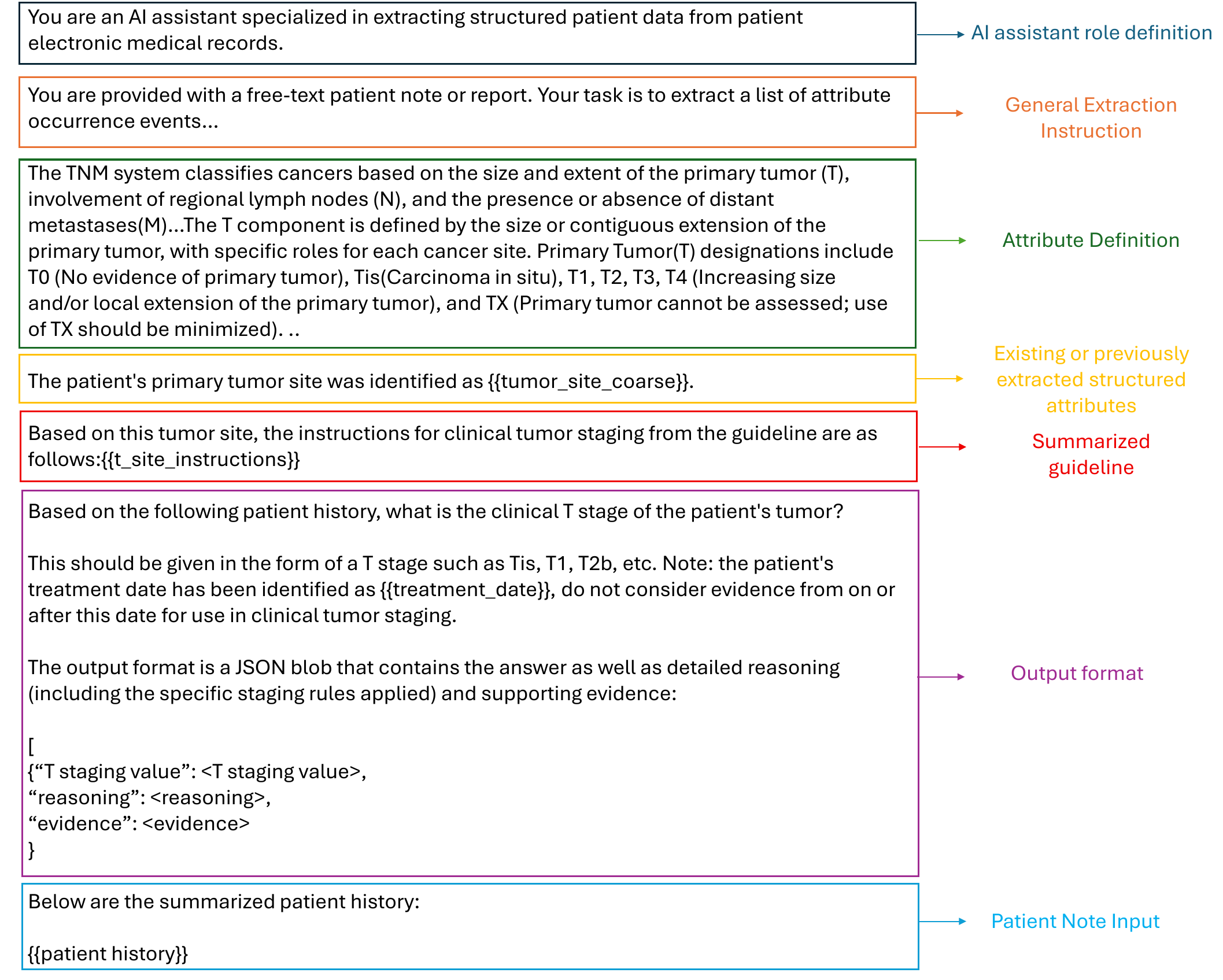}
\caption{Example long-context attribute \ours prompt for extracting clinical staging T from a history of patient notes \label{fig: example long context prompt}}
\end{figure}

\clearpage

\bibliography{iclr2024_conference}
\bibliographystyle{iclr2024_conference}

\clearpage

\clearpage

\section{Supplementary Material} 

{\renewcommand{\arraystretch}{1.5}
\begin{table}[!ht]
    \centering
    \caption*{Table S1. Oncology attributes and the associated event descriptors that can be defined as part of the attribute definition block in \ours prompt
    }
    \scalebox{0.6}{
        \begin{tabular}{|l|r|r|r|}
    \hline
    {\bf Attribute} & {\bf Descriptors} & {\bf Definition} & {\bf Extracted Example Values} \\ \hline
    
    \multirow{ 2}{*}{Case Finding} & cancer diagnosis & tumor histology & lung adenocarcinoma\\ \cline{2-4}
    & cancer diagnosis status & status of the diagnosis & positive, negative, suspicious, historical \\ \cline{2-4} 
         & date & cancer diagnosis date & 2016-12-15\\ \cline{2-4}
    \hline

    \multirow{ 2}{*}{PD-L1 Biomarker}
    & biomarker measurement type & specifies PD-L1 IHC measurement type & CPS, TPS, expression \\\cline{2-4}
    & biomarker variant & biomarker's variant or test value& 10\%, 5, High, T790M \\  
    \hline

    \multirow{ 3}{*}{Performance Status} & performance status & performance status snippet & ECOG 1, KPS 90\%  \\\cline{2-4}
    & performance status measurement type & the scale used for performance status & ECOG, KPS, Lansky \\\cline{2-4}
    & performance status value & the value of the performance status & CPS, TPS, expression \\ 
    \hline

    \multirow{2}{*}{Treatment} & treatment & treatment name & pembrolizumab, carboplatin, radiation \\\cline{2-4}
    & treatment date & treatment start date & 2014-02-03 \\
    \hline

    \multirow{ 3}{*}{ \begin{tabular}{@{}c@{}}Response \\ Progression \end{tabular}} & response & tumor response events & \begin{tabular}{@{}c@{}}complete response, partial response, \\ progressive disease, stable disease\end{tabular}  \\ \cline{2-4}
    & response disease & the disease or organ associated with the response & brain, lung, colon \\ 
    \cline{2-4}
    & response disease type & the type of disease with regard to cancer & tumor, lymph node, or non-cancerous tissues\\ 

    \hline
        Primary Site Coarse & Primary Site Coarse & Body site of primary tumor & C50 (Breast), C34 (Lung)\\
    \hline
        Primary Site Fine & Primary Site Coarse & Body site of primary tumor & C50.4 (Upper-outer quadrant of breast)\\ 
    \hline
        Histology & Histology & Cell type of tumor & 8046 (non-small cell lung cancer) \\
    \hline
        Clinical T & Clinical T & Clinical tumor staging & None, cT1, cT2, cT3, cT4 \\
    \hline
        Clinical N & Clinical N & Clinical nodal staging & None, cN0, cN1, cN2b \\
    \hline
        Clinical M & Clinical M & Clinical metastatic staging & None, cM0, cM1 \\
    \hline
        Pathologic T & Pathologic T & Pathologic tumor staging & None, pT1, pT2b \\
    \hline
        Pathologic N & Pathologic N & Pathologic nodal staging & None, pN0, pN1, pN2b \\
    \hline
        Pathologic M & Pathologic M & Pathologic metastatic staging & None, pM1 \\
        
\hline
    \end{tabular}
    }
    \label{tab:prompt-attribute-keys} 
\end{table}}

\begin{table}[!ht]
    \footnotesize
    \centering
        \caption*{Table S2. Description of the test sets from Providence. Path. = Pathology. For performance status, we include the following note types: Progress Notes + Telephone Encounter + H\&P + Consults + Discharge Summary + Assessment and Plan Notes + Plan of Care + Research Note + Treatment Plan}

     \scalebox{0.7}{
    \begin{tabular}{lrrrrr}
    \toprule
        {\bf Attributes} & {\bf \# patients} & {\bf \# notes} & {\bf \# Attribute Occurrences} & {\bf Note and Report Types} & {\bf Tumor Types} \\ 
      \midrule
        PD-L1 biomarker & 298 & 298 & 173 & Path. Reports + Progress Notes & All \\ 
        Performance Status & 45 & 565 & 45 & Progress Notes + Telephone Encounter...
        & All\\
        Treatment & 18 &431 & 203& Progress Notes & Lung Cancer \\
        Progression & 70 & 243 & 27 & Imaging Reports & Lung Cancer  \\
        Response & 70 & 243 & 28 & Imaging Reports & Lung Cancer \\
        Case Finding & 10,501 & 59,618 & 10,501 & Path. and Imaging \&Reports & All \\
       Long-context Attributes & 2,918 & 33,293 & 2,918 & Path., imaging and surgical reports & All\\
        
    \bottomrule
    \end{tabular}}
    \label{tab:note-level-dev-dataset}
\end{table}

\begin{table}[h]
\centering
\caption*{Table S3. Number of patients, notes and per-patient counts in the Providence and Hopkins datasets for long-context attribute abstraction}
\begin{tabular}{lr@{\hspace{3pt}}r}
\toprule
 & \multicolumn{1}{c}{Providence} & \multicolumn{1}{c}{Hopkins} \\
\midrule
\# of patients & 2,918 & 592 \\
\# of notes (all) & 33,293 & 7,555 \\
\hspace{5mm}\# of imaging reports & 21,936 & 6,221 \\
\hspace{5mm}\# of pathology reports & 11,357 & 1,334 \\
\hdashline
\# median per-patient tokens & 3984 & 5352\\
\# median per-patient notes & 7 & 10 \\
\# median per-patient imaging reports & 6 & 9\\
\# median per-patient pathology reports & 2 & 2 \\
\bottomrule
\end{tabular}

\label{tab:supple_prov_jh_dataset_comparison}
\end{table}

\subsection{Short-context Attribute Abstraction Tasks \label{appendix: short context attributes}}
We test \ours on six short-context attributes that can be understood and abstracted in the immediate context within a single note. 
\Cref{fig:uma_overview}B shows the template for abstracting short-context attributes. For the per-task configurations, we provide the attribute definitions according to \Cref{tab:prompt-attribute-keys}. 
For the per-patient input block, we input each note separately, and then collect the outputs for each patient for the final postprocessing step. For evaluation, we attach patient ID as an additional key to each attribute occurrence and count as positive when all the keys and values of the attribute correctly match the groundtruth. We report precision, recall and F1. We obtain manual annotations and relevant patient notes from Providence Health. 

\paragraph{PD-L1} PD-L1 protein expression is an important biomarker used to predict immunotherapy outcome as a high PD-L1 level may respond well to certain immune checkpoint inhibitor. Being able to extract the PD-L1 biomarker attribute can significantly facilitate the patient recruiting process in clinical trial matching. To configure the attribute definition block, we define the PD-L1 attribute by identifying two descriptors: the biomarker measurement types (eg. Combined positive score (CPS) or tumor proportion score (TPS)) and the biomarker variant descriptor that outputs the measurement values. To create the evaluation dataset, we manually curated 173 labels from 298 patients from the Providence data.

\paragraph{Performance Status} 
Performance status is a standard clinical criterion used to assess a patient’s ability to carry out daily activities and is commonly required for eligibility in cancer clinical trials.

To construct the evaluation dataset, we manually labeled performance status in clinical notes from 45 patients in the Providence dataset. A single clinician performed the initial annotations. Model predictions were generated using the o3 model, and any discrepancies between model output and annotations were adjudicated by a second clinician.

The model was instructed to extract ECOG scores either from explicit documentation (e.g., “ECOG 2”) or, when absent, to infer the score based on narrative descriptions of functional status (e.g., self-care, ambulation, bed confinement). When both an explicit score and a functional description were present but inconsistent, the inferred score based on functional status was prioritized. If the note lacked sufficient information, the model returned “N/A.”

\paragraph{Treatment}
The treatment attribute is a fundamental attribute of patient data, providing critical information about the timing and nature of treatments a patient has received. This data is essential for various downstream applications, such as predicting treatment outcomes and matching patients to clinical trials, which often require participants with specific prior treatments. In the task configuration section of the template, we define two key descriptors for the treatment attribute: the date of treatment and the treatment name. To create the evaluation set, we leverage existing treatment metadata from Providence. However, we observed that this structured data does not always capture all the treatments mentioned in the reports. As a result, we reviewed and manually corrected the data from a randomly selected subset to create a gold-standard test set.

\paragraph{Response and Progression}
Response and progression are important attributes to assess the treatment outcome of a clinical trial. In general, response indicates that the patient is showing improvement with the treatment, while progression signifies a worsening of the patient's condition. The response and progression attributes are extracted in one prompt. To provide the task-specific configurations in the template, we define response and progression based on the RECIST guideline\citep{nishino2010revised}\footnote{To curate the annotations from the reports, we relaxed the RECIST criteria to accommodate the level of details commonly available in standard follow-up radiology reports.}. Specifically, we require \ac{LLM} to list the specific response labels (choosing from: partial response, complete response, progressive disease or stable disease) and the corresponding response disease for each attribute occurrence. For each note, we collect the response attributes and progression attributes separately from the same \ac{LLM} output: if there is an occurrence of partial response or complete response for a disease in the note, we assign a response label. If there is an occurrence of progressive disease event in the note, we have a progression label. We manually curated the labels which are divided into 261 train labels and 55 test labels (28 response labels and 27 progression labels).

\paragraph{Case Finding}
Case finding is a system for locating patients who is diagnosed at a particular time. Case finding is essential for ensuring that cancer registries provide comprehensive, accurate, and timely data, which is critical for research, public health planning, and improving patient outcomes. To extract the case finding attribute, we define the diagnosis time as the key descriptor in the attribute definition block of our template as the ultimate goal of the task is to identify the moment of cancer diagnosis. We evaluate case finding extraction with labels collected from the cancer registry following the method in \citet{preston2023toward}. With around 50k \eat{50330} train labels, we provide a supervised baseline BERT model that predicts a binary label of whether the diagnosis happens given the note date following the setup in \citet{preston2023toward}.

\subsection{Long-context Attribute Abstraction: Cancer Staging \label{appendix: long context attribute}}

Cancer staging is a process used to determine the extent of cancer in the body. It involves the abstraction of multiple long-context attributes that requires a model to follow the rules and definitions set up in the lengthy clinical guidelines, and make inferences across multiple notes from the entire patient history.
For example, tumor measurements from imaging reports must be correlated with pathology findings to confirm a primary tumor site, and understanding the timing of diagnosis and treatments is crucial for determining their relevance to clinical or pathological staging values. \ours can effectively address these challenges. In the task configuration part, we offer the flexibility to take in long clinical guidelines and provide summarization specific to each attribute. These guidelines include the International Classification of Diseases for Oncology (ICD-O) manual, the American Joint Committee on Cancer (AJCC) Cancer Staging Manual, and the Standards for Oncology Registry Entry (STORE) manual. These sources were structured via a semi-automated process using GPT-4 to organize and summarize guidelines relevant to a particular abstraction task. To configure the attribute definition block, we define the attribute as the main descriptor to be extracted and since cancer staging involves complex reasoning and grounding, we also define two additional descriptors including the model reasoning descriptor and the evidence descriptor that enforce the \ac{LLM} to generate rationale and the supporting evidence (i.e. the piece of text from the patient note that supports the extracted attributes) before generating the attribute value. 
As to patient input, given the cancer staging attributes require reasoning over the patient history, we offer the solution to generate attribute-specific summaries from the patient history and pass the summaries into the patient input block. Alongside unstructured notes, we also provide the flexibility to incorporate other existing structured data such as treatment date. The previously extracted cancer staging attributes can also be chained for the best result. For example, we extract tumor site first and then we use the tumor site information to guide the extraction of the T/N/M attributes. \Cref{fig:uma_overview}C shows the full instantiation of \ours for this case.

In this study on cancer staging, we focus on abstracting the following eight attributes that are key in the staging process.

\paragraph{Primary site coarse/fine-grained}: The primary site attribute extracts the primary site of the tumor and we extract two attributes with different granularity. In the attribute definition block in our universal template, we define the primary site coarse attribute as the coarse-grained tumor site with a finite set of choices (eg. C34 for ``Lung''). For the fine-grained primary site attribute, we instruct the model to extract a more specific location (e.g. C34.1 for ``Upper lobe of the lung''). 

\paragraph{Histology} Histology describes the type of cells or tissues from which the cancer originates. We define the task requirements in the attribute definition block and instruct the model to output the four-digit ICD-O-3 histology code.

\paragraph {Clinical T/N/M} We follow the conventional TNM system in determining cancer staging \citep{rosen2023tnm}. We first define the clinical staging attributes which describe staging results determined before treatment initiation. Specifically, we define clinical T (tumor) attribute as describing the size and extent of the primary tumor and it is a multi class classification task.  Similarly, we define clinical N (nodule) attribute which describes the involvement of regional lymph nodes, and we define clinical M (Metatasis) as describing whether the cancer has metastasized. The exact definitions of the output values come from the attribute-specific summarization of the clinical guideline that lays out the detailed rules and requirements for determining the staging outcome. To prepare the patient input, we also pass in other structured data alongside the unstructured notes. For example, we pass in treatment date which is essential for guiding the \ac{LLM} to extract staging information only from records prior to the treatment date. We also input primary site attributes that we have previously extracted as additional structured data context to guide \ac{LLM} to provide the correct staging results.

\paragraph{Pathologic T/N/M} The pathologic T/N/M attributes have the same setup to clinical T/N/M apart from the different sources of staging information, allowing the use of staging information collected during treatment including excised tumor tissue.

\paragraph{Dataset and baselines}
Our primary dataset was obtained from Providence Healthcare where we collected electronic patient records and linked them to their cancer registry. We excluded cases lacking a pathology report within 30 days of the diagnosis date listed in the cancer registry. Additionally, patients with multiple primary cancer diagnoses were excluded from the study.

To provide a more comprehensive evaluation and test the generalization abilities of our approach, in addition to the data from Providence Healthcare, we collaborated with members of Johns Hopkins Medicine to compile a similar validation dataset from their cancer registry and patient records. This process involved collecting data from the cancer registry, integrating it with electronic patient records, and filtering based on the presence of a pathology report near the diagnosis date. 
Characteristics of the two datasets prepared for cancer staging evaluation are detailed in Table S3 and Figure S1.
We note that in the Hopkins dataset, a patient generally has more imaging reports and the notes are usually longer. In addition to differences in clinical notes, the staging datasets have different distributions of primary tumor sites, shown in Figure S1. While the Providence dataset is dominated by breast cancer, the Johns Hopkins dataset has a much higher portion of lung, blood, pancreatic, and brain cancer.

To establish an upper bound performance from conventional methods, we train a supervised baseline using the dataset (with 23438 patients' medical records as the train and dev set) and follow the methods described in \citet{preston2023toward}. We modify the original tasks to make the evaluation more realistic by (1) including a "None" category for each prediction, allowing the model to indicate there is not enough information to make a prediction; (2) predicting standard four-digit histology codes; and (3) using the standard nodal staging instead of simplified N0/N+ classification. 
For the Hopkins dataset, we only have the test labels and we directly run the supervised model trained on the Providence dataset. For \ours, we provide a zero-shot setting and we use the same prompt template for the two datasets except for passing in different clinical guidelines that are used in Providence and Hopkins.
As to the underlying \acp{LLM} for \ours, we use GPT-4 and GPT-4o as they are the most competitive \acp{LLM} as verified by the short-context attribute absraction experiments. We report accuracy rather than F1 for all the tasks, because each patient will have a prediction for each attribute, even if the prediction is \emph{None} (i.e. not enough information to determine).

\begin{table}[!ht]
    \scriptsize
    \centering
    \caption*{Table S4. Testing \ours on short-context attribute abstraction tasks in oncology. We show zero-shot \ours with GPT-4o achieves overall F1 improvement compared with the conventional baseline approaches. For performance status, we also experimented with reasoning models such as o1 and o3 and have found further gains. 
}
    \begin{tabular}{llrrr}
    \toprule
      {\bf Patient Attributes} & {\bf Approach} & {\bf Precision} & {\bf Recall } & {\bf F1 } \\ 
      \midrule
        PD-L1 biomarker & Heuristics & 97.5 & 89.6 & 93.4  \\ 
        
        & \ours~(GPT-4)  & $97.0\!\pm\!1.0$ & $95.3\!\pm\!1.2$  & $96.1\!\pm\!1.0$  \\
       & \ours~(GPT-4o) & $\mathbf{97.7\!\pm\!0.8}$& $\mathbf {98.8\!\pm\!0.6}$ & $\mathbf{98.3\!\pm\!0.6}$\\
         \midrule
        Performance Status & Heuristics & 53.3 & 53.3 & 53.3  \\ 
        & \ours~(GPT-4) & $\mathbf{74.1\!\pm\!2.6}$ & $\mathbf{74.1\!\pm\!2.6}$ & $\mathbf{74.1\!\pm\!2.6}$ \\ 
  & \ours~(GPT-4o) &  $\mathbf{73.3\!\pm\!1.0}$ & $\mathbf{71.1\!\pm\!0.0}$& $\mathbf{72.2\!\pm\!0.5}$ \\
  & \ours~(GPT-o1) &  $\mathbf{81.3\!\pm\!3.5}$ & $\mathbf{77.1\!\pm\!1.3}$& $\mathbf{79.1\!\pm\!2.4}$ \\
  & \ours~(GPT-o3) &  $\mathbf{78.9\!\pm\!3.1}$ & $\mathbf{78.4\!\pm\!3.1}$& $\mathbf{78.7\!\pm\!3.1}$ \\

        \midrule
         Treatment & Heuristics & 85.0 &  85.0& 85.0\\ 
        & \ours~(GPT-4)  &$83.4\!\pm\!4.1$  & $\mathbf{89.3\!\pm\!4.2}$ & $86.2\!\pm\!3.0$\\
        & \ours~(GPT-4o) & $\mathbf{87.8\!\pm\!2.0}$  & $86.8\!\pm\!2.5$  & $\mathbf{87.3\!\pm\!1.8}$  \\
       
         \midrule
          Progression & Supervised Model & {\bf 75.9} & 81.5 & 78.6 \\ 
        & \ours~(GPT-4)  & $67.3\!\pm\!5.6$  & $\mathbf{100\!\pm\!0.0}$  &$80.3\!\pm\!4.0$  \\
        & \ours~(GPT-4o)& $69.2\!\pm\!5.9$ & $\mathbf{100\!\pm\!0.0}$ &  $\mathbf{81.8\!\pm\!4.0}$  \\
         \midrule
          Response & Supervised model & 68.4  & {\bf 92.9}  & {\bf 78.8} \\
        & \ours~(GPT-4)  & $67.7\!\pm\!6.1$  &  $89.4\!\pm\!4.6$ &  $76.9\!\pm\!4.5$  \\
        &\ours~(GPT-4o) & $69.2\!\pm\!5.9$ & $89.0\!\pm\!4.5$ & $77.7\!\pm\!4.4$\\
         \midrule
          Case Finding & Supervised Baseline & 88.5 & {\bf 94.3} & {\bf 91.3} \\
        & \ours~(GPT-4) &  $86.9\!\pm\!0.3$ &  $89.8\!\pm\!0.3$ & $88.3\!\pm\!0.2$  \\
        & \ours~(GPT-4o) & $86.9\!\pm\!0.3$  & $89.9\!\pm\!0.3$ & $88.4\!\pm\!0.2$\\
        \midrule
        \midrule
          Average & SOTA baseline & 78.1	&82.8	&80.1
                   
                   \\

        & \ours~(GPT-4) &  $79.4\!\pm\!3.3$ 	&$\mathbf{89.7\!\pm\!2.2}$ &	$83.7\!\pm\!2.6$ 
 \\
        &{\bf \ours~(GPT-4o)} & $\mathbf{80.7\!\pm\!2.6}$ &	$89.3\!\pm\!1.3$ &	$\mathbf{84.1\!\pm\!1.8}$ \\
        \bottomrule
        
    \end{tabular}
    \label{tab:patient-structuring} 
\end{table}

\begin{table}
    \centering
    \caption*{Table S5. Performance comparison (measured in accuracy) of supervised models and zero-shot \ours on the long-context cancer staging attribute abstractions in the Providence dataset. The supervised baseline is trained with the Providence train set. }
    \begin{tabular}{lccc}
    \toprule
     & Supervised & \ours & \ours \\
     & (trained from Providence) & (GPT-4o) & (O1)\\
    \midrule
    Primary Site Coarse & $92.6\!\pm\!0.5$ & $93.2\!\pm\!0.5$ & $\mathbf{93.6\!\pm\!0.5}$ \\
    Primary Site Fine   & $\mathbf{71.8\!\pm\!0.9}$ & $68.6\!\pm\!0.9$ & $68.4\!\pm\!0.9$ \\
    Histology           & $\mathbf{80.9\!\pm\!0.8}$ & $74.3\!\pm\!0.8$ & $74.6\!\pm\!0.8$ \\
    Clinical T          & $\mathbf{58.7\!\pm\!1.0}$ & $51.3\!\pm\!1.0$ & $55.4\!\pm\!1.0$ \\
    Clinical N          & $91.0\!\pm\!0.6$ & $86.4\!\pm\!0.7$ & $\mathbf{91.2\!\pm\!0.5}$ \\
    Clinical M          & $\mathbf{95.2\!\pm\!0.4}$ & $92.3\!\pm\!0.5$ & $\mathbf{95.2\!\pm\!0.4}$ \\
    Pathologic T        & $55.0\!\pm\!1.0$ & $75.4\!\pm\!0.8$ & $\mathbf{77.5\!\pm\!0.8}$ \\
    Pathologic N        & $62.4\!\pm\!1.0$ & $75.8\!\pm\!0.8$ & $\mathbf{78.3\!\pm\!0.8}$ \\
    Pathologic M        & $93.8\!\pm\!0.5$ & $94.1\!\pm\!0.5$ & $\mathbf{96.0\!\pm\!0.4}$ \\
    \cmidrule(r){1-4}
    Average             & $77.9\!\pm\!0.3$ & $79.0\!\pm\!0.3$ & $\mathbf{81.1\!\pm\!0.3}$ \\
    \bottomrule
    \end{tabular}
        
    \label{table:prov_results}
\end{table}

\begin{table}[ht]  
\centering 
\caption*{Table S6. Performance comparison (measured in accuracy) of supervised models (trained with the labels from the Providence data) and zero-shot \ours on the long-context cancer staging attribute abstractions in the held-out Johns Hopkins Medicine dataset where no training labels are available.}

\begin{tabular}{lccc}
\toprule
 & Supervised & \ours & \ours \\
 & (trained from Providence) & (GPT-4o) & (O1) \\
\midrule
Primary Site Coarse & $82.6\!\pm\!1.5$ & $89.5\!\pm\!1.2$ & $\mathbf{90.9\!\pm\!1.2}$ \\
Primary Site Fine   & $67.5\!\pm\!1.9$ & $71.6\!\pm\!1.9$ & $\mathbf{72.6\!\pm\!1.8}$ \\
Histology           & $59.8\!\pm\!2.0$ & $\mathbf{61.2\!\pm\!2.0}$ & $60.7\!\pm\!2.0$ \\
Clinical T          & $49.8\!\pm\!2.1$ & $\mathbf{55.8\!\pm\!2.0}$ & $54.7\!\pm\!2.1$ \\
Clinical N          & $59.7\!\pm\!2.0$ & $68.0\!\pm\!2.0$ & $\mathbf{76.0\!\pm\!1.7}$ \\
Clinical M          & $83.0\!\pm\!1.6$ & $83.5\!\pm\!1.5$ & $\mathbf{89.1\!\pm\!1.3}$ \\
Pathologic T        & $73.2\!\pm\!1.8$ & $77.8\!\pm\!1.8$ & $\mathbf{83.4\!\pm\!1.5}$ \\
Pathologic N        & $52.7\!\pm\!2.0$ & $78.1\!\pm\!1.8$ & $\mathbf{85.4\!\pm\!1.4}$ \\
Pathologic M        & $71.4\!\pm\!1.9$ & $81.0\!\pm\!1.6$ & $\mathbf{87.0\!\pm\!1.4}$ \\
\cmidrule(r){1-4}
Average             & $66.6\!\pm\!0.8$ & $74.0\!\pm\!0.7$ & $\mathbf{77.7\!\pm\!0.7}$ \\
\bottomrule
\end{tabular}

\label{table:jh_results}
\end{table}

\begin{table}
\centering
\caption*{Table S7. Average per-patient token usage and cost using O1 on the Johns Hopkins dataset}

\begin{tabular}{lrrrr}
\toprule
 & Prompt & Completion & Reasoning & Cost (USD) \\
\midrule
Primary Site Coarse & 3\,104 & 2\,519 & 2\,267 & 0.20 \\
Primary Site Fine   & 2\,490 & 3\,288 & 2\,910 & 0.23 \\
Histology           & 4\,608 & 2\,777 & 2\,513 & 0.24 \\
Clinical T          & 10\,310 & 12\,911 & 11\,981 & 0.93 \\
Clinical N          & 2\,387 & 2\,359 & 2\,195 & 0.18 \\
Clinical M          & 2\,743 & 2\,851 & 2\,606 & 0.21 \\
Pathologic T        & 5\,826 & 5\,607 & 5\,001 & 0.42 \\
Pathologic N        & 3\,559 & 2\,567 & 2\,300 & 0.21 \\
Pathologic M        & 2\,926 & 2\,458 & 2\,153 & 0.19 \\
\midrule
\textbf{Sum}        & 37\,953 & 37\,336 & 33\,926 & 2.81 \\
\bottomrule
\end{tabular}
\label{table:Cost}
\end{table}

\begin{figure}[!ht]
\begin{AIbox}{Summarization Prompt}
{\fontsize{3}{7}\selectfont}
\begin{Verbatim}
  Briefly summarize the given clinical document, specifically
  including information related to cancer diagnosis and staging
  such as:

  - tumor information
      - detailed location
      - size
      - primary or metastatic
      - extent and structural invasion
  - tumor tissue characteristics
      - tumor morphology, including histology, behavior, 
        and grade
      - source, whether biopsy, resection, or cytology
  - lymph node involvement
  - explicit diagnosis or staging information
  - treatment information, such as chemotherapy, radiation, 
    or surgery

  The title of the summary should give information about 
  the procedure or observation being described.
  In particular, include the following information 
  for the specific note types below:
    Imaging Reports: include the imaging modality, 
    body site imaged, and purpose of the imaging
    Pathology Reports: include the type of tissue or fluid 
    being examined, including whether the tissue is 
    from a biopsy or resection
  For other note types, include the type and purpose of note.

  Additional summarization instructions:
    - Document the source of any tumor tissue, whether biopsy,
      resection, or cytology.
    - If there is no relevant information, generate a title 
      but leave the findings empty.

  The output format is a JSON blob that contains the summary 
  findings as well as evidence from the original note to 
  support each finding. The format is given below:

  {format_instruction}

  If there is no relevant information, create the title but 
  leave the findings as an empty list.

\end{Verbatim}
\end{AIbox}
\caption*{Figure S2. Prompt template for task-specific note-level summarization.}
\label{fig:summarization_prompt}
\end{figure}

\end{document}